\documentclass[5p,authoryear]{elsarticle}
\makeatletter 
\def\ps@pprintTitle{%
 \let\@oddhead\@empty
 \let\@evenhead\@empty
 \let\@evenfoot\@oddfoot} 
\makeatother
\usepackage[utf8]{inputenc} 
\usepackage[T1]{fontenc}
\usepackage[babel=true]{csquotes} 
\usepackage[fleqn]{amsmath} 
\usepackage{amsthm} 
\usepackage{booktabs} 
\usepackage{multirow} 
\usepackage{amssymb} 
\usepackage{float}
\usepackage{hyperref} 
\usepackage[french]{cleveref} 
\usepackage{algorithm}
\usepackage[noend]{algpseudocode}
\usepackage{mathrsfs}
\usepackage{mathtools}
\DeclarePairedDelimiter{\ceil}{\lceil}{\rceil}
\usepackage{graphicx}
\usepackage{pdflscape,array,booktabs}
\usepackage{threeparttable,booktabs,multirow,lscape}

\begin{document}
\begin{frontmatter}
\title{On the Current State of Research in Explaining Ensemble Performance Using Margins}

\author[add1]{Waldyn Martinez}
\ead{martinwg@miamioh.edu}
\author[add2] {J. Brian Gray}
\address[add1]{Department of Information Systems and Analytics, \\Miami University, Oxford, OH 45056, USA}
\address[add2]{Department of Information Systems, Statistics and Management Science\\ University of Alabama, Tuscaloosa, AL 25205, USA\\}
\cortext[add1]{Corresponding author}

\begin{abstract}
Empirical evidence shows that ensembles, such as bagging, boosting, random and rotation forests, generally perform better in terms of their generalization error than individual classifiers. To explain this performance, ~\citet{schapire98} developed an upper bound on the  generalization error of an ensemble based on the margins of the training data, from which it was concluded that larger margins should lead to lower generalization error, everything else being equal. Many other researchers have backed this assumption and presented tighter bounds on the generalization error based on either the margins or functions of the margins.  For instance, \cite {shen2010boosting} provide evidence suggesting that the generalization error of a voting classifier might be reduced by increasing the mean and decreasing the variance of the margins. In this article we propose several techniques and empirically test whether the current state of research in explaining ensemble performance holds. We evaluate the proposed methods through experiments with real and simulated data sets.
\end{abstract}


\end{frontmatter}

\section{Introduction}
\label{intro}
Boosting \citep {freund97} and other ensemble methods, such as bagging \citep{breiman1996bagging}, Random Forests  \citep{breiman2001random} and rotation forests \citep{rodriguez2006rotation}, create a set of weak classifiers from a base learning algorithm $\textbf{B}$, which are typically decision trees, then combine the predictions from the classifiers in the form of a weighted vote, to produce an improved prediction compared to individual classifiers \citep{drucker1994boosting, dietterich2000ensemble, breiman2001random, maclin2011popular}. Upper bounds based on the sample margins of the ensemble provide some explanation on why ensembles perform as well as they do. \cite {schapire98} first pointed to margins as a key determinant of ensemble performance. This assertion has led researchers to consider directly optimizing the margins or functions of the margins of the ensembles to improve their generalization accuracy  (see, e.g., \citealt{grove98, mason2000improved, shen2010boosting}). Several authors have concluded that maximizing the average margin, or maximizing the whole margin distribution should result in improved performance, but that simply maximizing the minimum margin generally does not contribute to better accuracy; in fact the opposite is more often true \citep{schapire1999theoretical, reyzin2006boosting, grove98, zhou2014large}. \cite{shen2010boosting}, for instance, provide arguments to suggest that the generalization error of combined classifiers might be reduced by increasing the mean and decreasing the variance of the margins, which we call here ``squeezing the margins". Other authors suggest that specific margin instances hold a clue to better generalization \citep{shen2010boosting, wang2011refined, wang2012further}. In this article, we design algorithms to empirically test whether the state of research in the explanation of ensemble performance translates into better performing algorithms. We do not question the theoretical soundness of the generalization error bounds, but simply test whether evidence suggests that better performing ensemble algorithms can be derived from the practical interpretations of the bounds. In the next section we discuss margins, the generalization error bounds based on the margins, and the large margins theory. In sections \ref{maxmarg} and \ref{maxvar}, we propose algorithms specifically designed to test the main research hypotheses presented in the bounds. We also show the results of carefully designed experiments using the proposed methods. Finally, we discuss our results and their implications in section \ref{conc}.

\section{Preliminaries}
\label{prelim}
We assume a set of $T$ classifiers, \(h_t(\textbf{x}), t = 1,2,...,T\), is created from a base learning algorithm $\textbf{B}$, each of which takes a \(p\times1\) input vector \(\textbf{x}\) and produces a prediction \(h_t(\textbf{x})\in \{-1,+1\}\) for a binary response variable \textit{Y}. The prediction can be extended to the multiclass case, but we focus on the binary framework in this paper. The combined classifier prediction $f(\textbf{x})$ of covariate vector $\textbf{x}$ is given by the sign of the linear combination of the $T$ individual classifiers:\\ 
\begin{equation}
\label{eq:1}
f(\textbf{x}) = sign \left(\sum_{t=1}^T \alpha_t h_t(\textbf{x})\right),
\end{equation}
where $\alpha_t$ is a weight associated with the $t^{th}$ weak classifier, $0\leq \alpha_t \leq 1$ and $\sum_{t=1}^T \alpha_t = 1$. The task of an ensemble algorithm is to create a set of weak learners and determine a set of weights $\{\alpha_1, \alpha_2, ..., \alpha_T\}$ based on a training sample of data, $\{( \textbf{x}_i, y_i ), i = 1,2, ...,n \}$, to produce a combined prediction with small generalization error.
The margin of the $i^{th}$ training observation is given by
\begin{equation}
\label{eq:2}
m_i =  y_i \sum_{t=1}^T \alpha_{t} h_{t}(\textbf{x}_i).
\end{equation}
The margin is a measure of the ``confidence" of the prediction for the $i^{th}$ training observation and is equal to the difference in the weighted proportion of weak classifiers correctly predicting the $i^{th}$ observation and the weighted proportion of weak classifiers incorrectly predicting the $i^{th}$ observation, so that $-1\leq m_i \leq 1$. A margin value of $-1$ indicates that all of the weak learner predictions were incorrect, while a margin value of $+1$ indicates all of the weak learners correctly predicted the observation. Let \(h_{it}=\pm1\) denote the prediction of the $t^{th}$ weak learner for the $i^{th}$ observation in the training data. We define the matrix
\begin{equation}
\label{eq:3}
H_{n,T} = \begin{pmatrix}
h_{11} & h_{12} & \cdots & h_{1T} \\
h_{21} & h_{22} & \cdots & h_{2T} \\
\vdots  & \vdots  & \ddots & \vdots  \\
h_{n1} & h_{n2} & \cdots & h_{nT}
\end{pmatrix},
\end{equation}
\noindent where $H \in \{-1,1\}^{n \times T}$ , to be the matrix of predictions for the $T$ weak classifiers. Boosting is one of the most well-known ensemble methods. The development of boosting algorithms was based on the PAC learning theory \citep{valiant1984theory}. The idea of boosting is based on the question posed by \cite{kearns1994cryptographic} on whether weak and strong learning are equivalent for efficient learning algorithms. In the development of boosting,  \cite{schapire1990strength} presented the first algorithm that transformed (``boosted") a weak learner into a stronger performing algorithm. Boosting is therefore not necessarily a single algorithm, but instead a family of algorithms with the strong PAC learning property. The strong PAC learning property states that for every distribution $P_{XY}$, all concepts $f \in \mathscr{F}$  and all $\epsilon \in (0,1/2)$, $\delta \in(0,1/2)$, a strong PAC learner has the property that with probability at least $1-\delta$, the base learning algorithm $\textbf{B}$ outputs a hypothesis $h$ with $P\left[h(\textbf{x})\neq f(\textbf{x})\right] \leq \epsilon$. $\textbf{B}$ must run in polynomial time in $1/\epsilon$, and $1/\delta$ using only a polynomial (in $1/\epsilon$ and $1/\delta$) number of examples. AdaBoost \citep{freund97} is the leading boosting algorithm and has been shown to be a PAC (strong) learner. The AdaBoost algorithm is presented in Algorithm \ref{AdaBoost}.
\begin{algorithm}
	\caption{AdaBoost (AB)}
	\label{AdaBoost}
	\begin{algorithmic}[1]
		\State $[Input]: S =\{(\textbf{x}_i,y_i), i= 1,...,n\}$ for $T$ iterations
		\State $[Initialize]: D_i^{(1)} = \frac{1}{n}$
		\State $[Loop]:$ Do For $t = 1,...,T$
		\State \indent (a) Train $h_t$ on the sample set $\{S, D^{(t)}\}$
		\State \indent (b) Set $\epsilon_t = \sum_{i=1}^{n} D_i^{(t)}I \left(y_i \neq h_t(\textbf{x}_i)\right)$
		\State \indent (c) Break if  $\epsilon_t = 0$ or $\epsilon_t \geq \frac{1}{2}$
		\State \indent (d) Set  $\alpha_t = \frac{1}{2} \ln \left(\frac{1-\epsilon_t}{\epsilon_t}\right)$ 
		\State \indent (e) Update  $D_i^{(t+1)} = \frac{ D_i^{(t)}\exp\{-\alpha_ty_ih_{t}(\textbf{x}_i)\}}{\sum_{i=1}^n D_i^{(t)}\exp\{-\alpha_ty_ih_{t}(\textbf{x}_i)\}}$
		\State $[Output]: f{\textbf{(x)}}= sign  \left(\sum_{t=1}^T\alpha_th_t(\textbf{x})\right)$.
	\end{algorithmic}
\end{algorithm}

Bagging, short for bootstrap aggregation, is another strong performing ensemble method for combining several base learners to produce a more accurate prediction. Given a training set of size $n$, bagging uses bootstrapping to generate a new training set  of size $n$  and fits a weak learner to the data. This process is repeated $T$  times, and the final classification aggregation can be a majority vote for the classification problem, or an average of the predicted values for regression problems. Bagging improves the performance of base classifiers, especially for unstable learners that vary significantly with small perturbations of the data set, e.g., decision trees. \cite{breiman1996bagging} suggested that the variance, which was defined as the scatter in the predictions obtained by using different training sets drawn from the same distribution, was reduced in the combination created by bagging, classifying it as a variance-reducing ensemble algorithm.

Random Forests \citep{breiman2001random}  is also one of the strongest performing ensembles. A Random Forest (RF) is defined as a classifier consisting of a collection of trees  ${h_t(\textbf{x},\theta_t),t=1,...,T}$, where ${\theta_t}$ are independently and identically distributed random vectors. Each tree casts a unit vote for the most popular class at input $\textbf{x}$. RFs inject randomness by growing each of the $T$ trees on a random subsample of the training data, and also by using a small random subset of the predictors at each split decision. The RF method is similar to boosting in the fact that it combines classifiers that have been trained on a subset sample or a weighted subset, but they differ in the fact that boosting gives different weight to the base learners based on their accuracy, while RF classifiers have uniform weights. A general RF classification algorithm is presented in Algorithm \ref{RF}.

\begin{algorithm}
	\caption{Random Forest (RF)}
	\label{RF}
	\begin{algorithmic}[1]
		\State $[Input]: S =\{(\textbf{x}_i,y_i), i= 1,...,n\}$ for $T$ iterations
		\State $[Loop]:$ Do For $t = 1,...,T$
		\State \indent (a) Draw a bootstrap sample set $S_Z$ from $S$
		\State \indent (b) Draw a sample $p_Z$ of variables from $p$ 
		\State \indent (c) Train $h_t$ on the sample set $\{S_Z, p_Z\}$
		\State $[Output]: f{\textbf{(x)}}= \underset{y \in Y}{\operatorname{argmax}}  \sum_{t=1}^T\ I \left(h_t(\textbf{x})=y\right)$.
	\end{algorithmic}
\end{algorithm}

Many other ensemble methods have been proposed in the literature that have either completely original weak learner combination approaches or are modifications of the AdaBoost and Random Forests algorithms. Among them, we have LogitBoost \citep{friedman2000additive}, which uses maximum likelihood to minimize a logistic loss function, instead of that used in Algorithm \ref{AdaBoost}d.  MadaBoost \citep{domingo2000madaboost} and the methods presented in \cite{martinez2016noise} attempt to alleviate AdaBoost's lack of robustness in the presence of outliers and noise. Gradient boosting \citep{friedman2001greedy} and stochastic gradient boosting \citep{friedman2002stochastic} also use modifications of the loss function in Algorithm \ref{AdaBoost}d to provide robustness and approximation accuracy to the boosting algorithm. Other methods, such as Rotation Forests \citep{rodriguez2006rotation} and the ideas presented in \cite{zhang2009novel} use principal component analysis (PCA) applied to each round $T$ to produce the most diverse and best performing set of weak learners within the ensemble. Ensembles have also been developed using other popular techniques, such as Bayesian methods \citep{denison2001boosting, chipman2007bayesian, chipman2010bart}. This list of ensembles is by no means exhaustive, but it represents some of the most common algorithms used to date. There has been ample research on the various ensemble methods presented here and how they perform under different settings. For a more in-depth treatment on ensemble methods and their performance, the interested reader is referred to \cite{quinlan1996bagging, maclin1997empirical, opitz1999popular, dietterich2000ensemble, maclin2011popular}; and \cite{zhou2012ensemble}. 

Ensembles generally perform better than individual classifiers. To explain the superior performance of AdaBoost, \cite{freund97} suggested a bound on the generalization error of any ensemble  in terms of the number of classifiers combined, the VC-dimension (a measure of complexity) and the training set error rate. The bound is given in Theorem 1. \\

\noindent
{\bf Theorem 1 \label{thm1}}{\citep{schapire98}. Assuming that the base-classifier space $\mathscr{H}$ is finite, and for any  $\delta> 0$ and $\theta>0$, then with probability at least $1-\delta$ over the training set $S$ with size $n$, every voting classifier $f$ satisfies the following bound:} 
\begin{equation}
\label{bound:1}
P\left[f(\textbf{x})\neq y\right] \leq \hat{P}\left[m(\textbf{x},y) \leq \theta\right] + O\left( \sqrt{\frac{d} {n\theta^2}}{}\right),
\end{equation}
where $P\left[f(\textbf{x})\neq y\right] $ is the generalization error of the combined classifier. The term $\hat{P}\left[m(\textbf{x},y) \leq \theta\right]$ in (\ref{bound:1}) is the proportion of training set margins less than a value $\theta > 0$, and $d$ is the VC-dimension of the space of all possible weak classifiers (a measure of complexity).  \cite{freund97} used this bound to provide an explanation for the good performance of boosting, which they show is highly effective in increasing margins. Researchers have used the bound in  (\ref{bound:1})  as a means to improve upon the explanation of ensemble methods through VC-type bounds and to also conclude that higher margins should lead to a lower generalization error rate, everything else being equal. (See, e.g., \citealt{schapire98, grove98, mason2000improved, reyzin2006boosting, shen2010boosting, wang2011refined, cid2012three, martinez2014role, liu2015boosting}; and \citealt{zhang2016optimal} for statements about the importance of large margins.)  The phrase ``maximizing the margins" has been used extensively in the literature, however there is no operational guideline  on what it means. For instance, \cite{grove98} and others have defined ``maximizing the margins" as maximizing the minimum margin. Their linear programming approach, LP-Boost, is designed to maximize the minimum margin by optimizing the weights associated with the weak learners.  \cite{breiman1999prediction} presented a bound on the generalization error based on the minimum margin that was tighter than that presented by \cite{schapire98} in (\ref{bound:1}). The bound is shown in Theorem 2.\\

\noindent
{\bf Theorem 2 \label{thm2}}{\citep{breiman1999prediction}. Let $\theta_0=\min(m_i)$ be the minimum margin, and if $\theta_0> 4 \sqrt{\frac{2}{\left|\mathscr{H}\right|}}$, and a given value $R = 32 \ln {\frac{2\left|\mathscr{H}\right|}{n\theta_0^2}}\leq 2n$. Assuming that the base-classifier space $\mathscr{H}$ is finite, that is $\left|\mathscr{H}\right|<\infty$, and  for any $\delta> 0$, then with probability at least $1-\delta$ over the training set $S$ with size $n$, every voting classifier $f$ satisfies the following bound:} 
\begin{equation}
\label{bound:2}
P\left[f(\textbf{x})\neq y \right]\leq R \left( 1+\ln{2n}+ \ln{\frac{1}{R}}\right)+ \frac{1}{n} \ln{\frac{\left|\mathscr{H}\right|}{\delta}}.
\end{equation}
\cite{breiman1999prediction} also produced an algorithm, called arc-gv, that explictly maximized the bound in (\ref{bound:2}). \cite{breiman1999prediction} found in his experiments that arc-gv not only produced larger minimum margins over all training examples, but it also produced a better margin distribution than AdaBoost, yet his algorithm more often than not performed worse than AdaBoost in terms of the generalization error. \cite{reyzin2006boosting} replicated the analysis in \cite{breiman1999prediction}, but pointed out that the trees (weak learners) found by arc-gv were deeper on average than the trees (weak learners) explored by AdaBoost, even though the number of terminal nodes was kept the same. They concluded that this increased complexity could have led to overfitting by arc-gv, and hence the worse test set performance, but more importantly, that the increased complexity violated the assumption of everything else being the same, which discredited Breiman's evidence against the large margins theory. 

Next, we will present the most recent significant work in the margins explanation of ensemble performance, and we will also propose algorithms to empirically test the hypotheses proposed in these works.

\section{Maximizing the Margin Distribution}
\label{maxmarg}
After the less-than-satisfactory results from maximizing the minimum margin, many authors have proposed optimizing other functions of the margin distribution. For instance, \cite{reyzin2006boosting} suggested maximizing the average or the median margin, while \cite{mason2000improved} proposed the DOOM (Direct Optimization of Margins) algorithm, which optimizes the average of a cost function of the margins. DOOM outperforms AdaBoost in many of the experiments considered. \cite{mason2000improved} also confirm that the size of the minimum margin is not a critical factor in generalization performance of an ensemble solution. \cite{shen2010boosting} suggested that AdaBoost inherently attempts to maximize the average margin while minimizing the variance of the margin distribution.  The results in \cite{shen2010boosting} are summarized in the following assertion:\\

\noindent
{\bf Theorem 3 \label{thm3}}{\citep{shen2010boosting}.  ``AdaBoost maximizes the unnormalized average margin and simultaneously minimizes the variance of the margin distribution under the assumption that the margin follows a Gaussian distribution."\\
	
	\begin{table*}[t]
		\centering
		\caption{Description of Data Sets}
		\begin{tabular}{lllccc}
			\toprule
			Data Set & Description &  Source & Training & Testing  & Features  \\
			\midrule
			Australian & Australian Credit Approval&\cite{Lichman2013} &690 &         & 14      \\
			Breast Cancer & Breast Cancer Wisconsin & \cite{Lichman2013}  &683   &       & 10    \\
			Colon Cancer & Colon Cancer Data & \cite{alon1999broad}  &62   &       & 2000    \\
			Diabetes & Diabetes Patient Records & \cite{Lichman2013} &768   &       & 8     \\
			Four Class & Fourclass Non-Separable  & \cite{ho1996building} &862   &       & 2     \\
			Ionosphere & Ionosphere Data Set & \cite{Lichman2013}  &351 &  & 34    \\
			Madelon & Artificial Data & \cite{guyon2004result}  &2000  & 600   & 500   \\
			Mushrooms & Mushrooms Data Set& \cite{Lichman2013}  &8124  &    & 112   \\
			Musk & Molecules Prediction& \cite{Lichman2013}  &6598  &    & 168  \\
			Parkinsons & Parkinsons Disease& \cite{Lichman2013}  &197  &    & 23  \\
			Pima& Pima Indians Data Set& \cite{Lichman2013}  &768  &    & 8   \\
			Sonar  & Sonar Data Set  & \cite{Lichman2013} &208  &   & 60    \\
			Spambase& Spam Emails Data Set& \cite{Lichman2013}  &4601  &    & 57   \\
			Splice  & DNA Splice Junctions   & \cite{Lichman2013} &1000  & 2175  & 60   \\	
			Transfusion	& Blood Transfusion Data Set& \cite{Lichman2013}  &748 &    & 5   \\
			\bottomrule
		\end{tabular}%
		\label{tab:addlabel}%
	\end{table*}%
	
	\noindent \cite{shen2010boosting}  then propose an algorithm named MD-Boost (Margin Distribution Boosting) that maximizes the average margin while reducing the variance of the margin distribution. \cite{shen2010boosting} also provide evidence that MD-Boost outperforms AdaBoost in many of the experiments using UCI-Repository data sets. \citet{germain2015risk} theoretically analyze the relationship between the variance of the margins and the risk of majority voters. To be more precise, \citet{germain2015risk} bounded the risk of a classifier with the expected disagreement between the individual learners. \citet{germain2015risk} define $M(\textbf{x},y)$ as a random variable, that given an example $(\textbf{x},y)$ drawn according to $D$, outputs the margin of the majority voter on that example. The generalization error, or risk of $f$ can therefore be defined in terms of the margins, as the probability that the majority voter is incorrect $R[f] = P\left[M(\textbf{x},y) \leq 0 \right]$.  The second moment $\mu^2_M = E_{D}M^2(\textbf{x},y)$ of the distribution of $M(\textbf{x},y)$ is of particular importance. \citet{germain2015risk} provide an upper bound on the generalization error $R[f]$ of an ensemble that relates the diversity or expected disagreement between voters $(d_S)$, which is a particular measure of diversity of a voting classifier $div(f)$ and the second moment of the margin distribution $\mu^2_M$. The bound is given in Theorem 4. \\
	
	\noindent
	{\bf Theorem 4 \label{thm3}}{\citep{germain2015risk}. For any distribution $Q$ on a set of voters and any distribution $D$ on $X$, if $\mu_M>0$, we have:}
	
	\begin{equation}
	\label{bound:3}
	P\left[f(\textbf{x})\neq y\right] \leq 1 - \frac{1-2R_D(G_Q)}{1-2d^D_Q},
	\end{equation}
	where $R_D$ is the Gibbs risk of the classifier and $1-2d^D_Q$ relates the risk of the classifier to the second moment of the margin distribution. The reader is referred to \citet{germain2015risk} for a more complete explanation of the relationship and the derivation of the bound. We can also conclude from the bound in (\ref{bound:3}) that reducing the second moment $\mu^2_M$ of the margins of any given ensemble should produce a more diverse and better performing ensemble classifier. Theorems 3 and 4 suggest that reducing the variation of the margins, while increasing the mean of the margin distribution, might result in better performing ensembles, holding all other factors constant, such as the complexity of the base learning algorithm $\textbf{B}$. 
	
	\subsection{Weight-Based Margin Optimization}
	The methods described here aim to improve the margins of any ensemble, including those of AdaBoost and Random Forests, by simply optimizing the weights of the weak learners of the given ensemble solution. Researchers have previously used linear and quadratic programming to optimize the weights of a given ensemble solution in an effort to improve upon its performance (see, e.g., \citealt{grove98, ratsch2001robust, ratsch2002maximizing, ratsch2005efficient, wang2011refined}). The proposed methods use linear programming (LP) to improve upon the whole margin distribution of an ensemble solution, but are also tailored to increase specific margin instances. Figure \ref{fig:1} illustrates the changes in the margin distributions as the number of weak learners increase. It is evident that $T$ grows larger, the lower margins shift to the right, and the variance appears to reduce. AdaBoost is especially aggressive in increasing the lower margins percentiles.  The task of our proposed methods is to determine a new set of weights $\{w_1, w_2, ..., w_T\}$ from the given ensemble solution $h_t(\textbf{x}), t = 1,2,...,T$, $0\leq w_t \leq 1$ and $\sum_{t=1}^T w_t = 1$, such that there is improvement in the margin distribution in agreement with given hypothesis we want to test. Note that for the training sample of data $\{( \textbf{x}_i, y_i ), i = 1,2, ...,n \}$ used to produce the ensemble solution, the values of the $T$ weak learner predictions $h_{it}$ for each observation $i$  and the original solution weights $\{\alpha_1, \alpha_2, ..., \alpha_T\}$ are fixed. To avoid other factors influencing the generalization error bounds, we fix the complexity of the ensemble solutions by using the same set of trees generated by the ensemble, and also by forcing the trees to grow to $k = 4$ terminal nodes with a fixed depth of 2. We should mention that our simulation results continue to hold regardless of the selected value of $k$, including trees ranging from decision stumps $(k = 2)$ to unpruned trees.
	\begin{figure}[h]
		\centering
		\includegraphics[width=8cm,height=6.5cm]{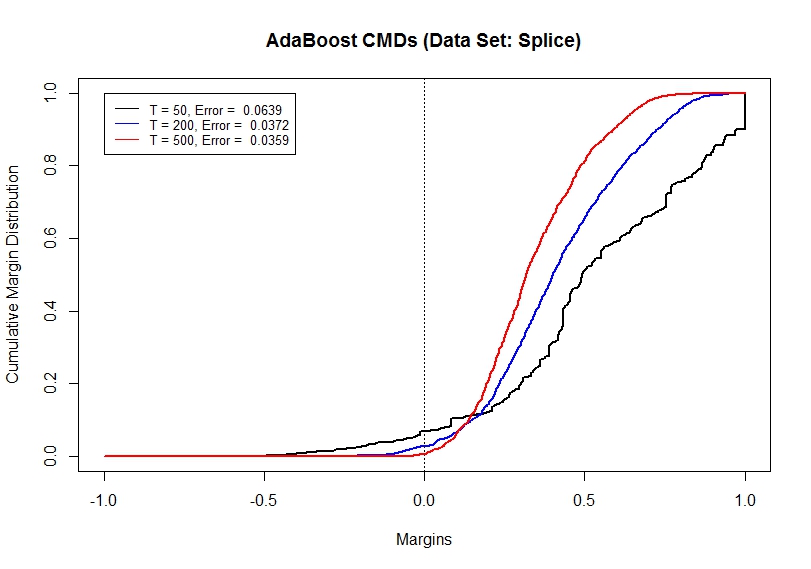}
		\caption{AdaBoost cumulative margin distributions (CMDs) for $T=\{50, 200, 500\}$ for the Splice data set using 4-node decision trees.}
		\label{fig:1}
	\end{figure} 
	In Algorithm \ref{MM} we show a linear programming (LP) optimization algorithm that can place emphasis on the improvement of specific margins for a given ensemble solution, that is, we weight the importance of margin improvements to give more emphasis to either all the margins or the lower percentile margins by utilizing a fixed set of weights $r_i,  i = 1,2, ...,n$ on top of the maximization of the sum of the margin improvements. The linear program formulation is shown in Algorithm \ref{MM}.
	
	\begin{algorithm}
		\caption{MM Algorithm}
		\label{MM}
		\[ \begin{array}{rl}
		\max &  \sum{_{i=1}^nr_i \left[y_i\sum{_{t=1}^Tw_th_{it}-m_i}\right]}  \\
		\mbox{s.t.} & y_i \sum{_{t=1}^T w_t h_{it} \geq m_i}, i = 1,2,..., n \\
		& \sum{_{t=1}^T w_t= 1}\\
		& w_t \geq 0, t =1,2,..., T\\
		\end{array}
		\]
	\end{algorithm} 
	\noindent The value $m_i = \sum{_{t=1}^T y_i h_{it} \alpha_t}$ is the original margin instance for the $i^{th}$ observation and $w_t , t =1,2,..., T$ are the optimized weights for the weak learners generated by the LP. The constraint $ y_i \sum{_{t=1}^T w_t h_{it} \geq m_i}$ forces the new margin for each instance to be as large as the original margin. Various options exist for how to select these $r_i$  weights to focus on specific margins. We present different weighting schemes depending on the specific purpose. 
	
	
	\subsection{Margin Maximization  (Margins Weighted Equally)}
	\label{MM_UWS_SECTION}
	\begin{figure*}
		\includegraphics[width=\textwidth]{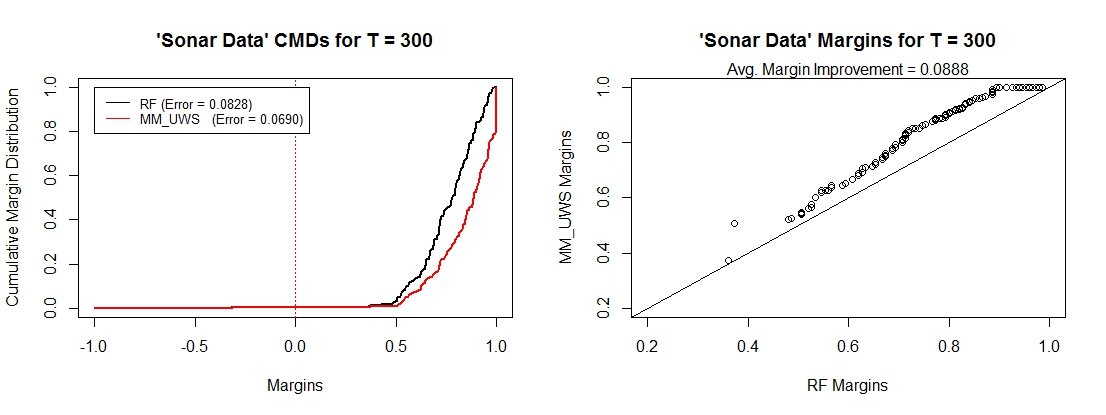}
		\caption{Comparisons of RF and the Uniform Weighting Scheme (MM\_UWS) for the Sonar Data using 300 weak learners. The left panel compares the cumulative margin distributions (CMDs) for RF and the MM\_UWS algorithm. The right panel is a scatter plot of the AdaBoost margins and MM\_UWS margins. The line in the graph indicates equality of the margins.}
		\label{fig:2}       
	\end{figure*}
	
	We develop here an algorithm to maximize the margin distribution by weighting all margins equally. To achieve that, we modify Algorithm \ref{MM} and use the weighting scheme $r_i = 1$, which results in maximizing the sum of the margin improvements. We call this formulation the MM\_UWS (Uniform Weighting Scheme) algorithm. All of the margins are weighted equally using this scheme. With this weighting scheme we test the hypothesis proposed in \cite{schapire98}, which states that holding all other factors constant, an improvement (increase) on the margins should result in a better performing ensemble. We are also testing whether maximixing the average margin is crucial to an improved performance \citep{reyzin2006boosting}. Figure \ref{fig:2} illustrates the optimized margins of the MM\_UWS method versus the original margins produced by a Random Forest solution.  We also plot the two cumulative margin distributions (CMDs) for comparison. The example, which is based on a 300-tree ($k = 4$, depth = 2) Random Forest ensemble constructed on the Sonar data set using a 70/30 train and validation sampling scheme, shows that the entire margin distribution of the RF ensemble was improved. The minimum margin improvement for this particular example was 0.0133, while the average improvement was 0.0888. In other words, all of the original margins provided by the RF solution were improved by at least 0.0133. The performance of the MM\_UWS algorithm was comparable to that of the RF solution in terms of the test set error rate. The MM\_UWS algorithm achieved a 0.0690 test error rate, compared to a  0.0828 for the RF solution. In this case the performance of the MM\_UWS was better, however this is not always the case.

	To better gauge the performance of the MM\_UWS algorithm in different situations, we perform experiments with 15 real and simulated data sets using Random Forests and AdaBoost. Table \ref{tab:2} illustrates the performance of the MM\_UWS algorithm versus a Random Forest of 200 decision trees, repeated for 100 simulations. The results here extend to other ensemble sizes and tree topologies. We can see in Table \ref{tab:2} that an improvement in the mean of the margins $\bar{m}$  or an improvement (increase) in the whole margin distribution does not necessarily improve the generalization performance, despite holding the trees to the same complexity as those used by the Random Forests solution. In fact, the opposite is most often true. A similar story can be seen in Table \ref{tab:3}, where the results of 100 simulations of the MM\_UWS algorithm are compared against an AdaBoost solution of 200 decision trees. AdaBoost performs better in spite of improvements to the margin distribution of the ensemble. We find that the MM\_UWS algorithm is able to find optimized weights $w_t$ that produce larger margins over all training examples and conclude that does not necessarily result in a improved generalization performance of the given ensemble solution. This supports the findings in \cite{breiman1999prediction} and contradicts the main hypothesis proposed in \cite{schapire98} and reinforced by many other researchers.
	
	\begin{table}
		\centering
		\caption{MM\_UWS vs Random Forests ensemble of 200 (depth = 2, forced to 4 terminal nodes) CART trees for 100 simulations. (*) in the test error indicates statistically significant ($\alpha = 0.05$) better performance on a paired t-test.}
		\begin{tabular}{lcccc}
			\toprule
			& \multicolumn{2}{c} {Test Error} & \multicolumn{2}{c} {Margin Improve} \\ [0.5ex] 
			Data Set & RF & UWS  & Mean & Min  \\
			\midrule
			Australian &0.0565 & 0.0567 & 0.0152  & 0.0000\\
			BreastCancer &0.0121* & 0.0144 & 0.0201 & 0.0000\\
			ColonCancer &0.0786* &  0.1104 &  0.2026  & 0.0186\\
			Diabetes  &0.1039  & 0.1039 & 0.0000 & 0.0000\\
			Four Class  &0.0022*  & 0.0033 & 0.0289 & 0.0000\\
			Ionosphere  &0.0289 & 0.0300 & 0.0449  & 0.0000\\
			Madelon  &0.0827 & 0.0827 & 0.0000  & 0.0000\\
			Mushrooms &0.0000*  & 0.0035 & 0.0004  & 0.0000\\
			Musk &0.0485 & 0.0469* & 0.0311 & 0.0004\\
			Parkinsons &0.0459 & 0.0472 & 0.0809 & 0.0000\\
			Pima&0.1014 & 0.1014& 0.0000 & 0.0000\\
			Sonar &0.0773*  & 0.0847 & 0.0802 & 0.0065\\
			Spambase & 0.0211 &0.0211 & 0.0000& 0.0000\\
			Splice   &0.0661& 0.0661 & 0.0000  & 0.0000\\	
			Transfusion&0.1031* &  0.1038 & 0.0158& 0.0000\\
			\bottomrule
		\end{tabular}%
		\label{tab:2}%
	\end{table}%
	
	\begin{table}
		\centering
		\caption{MM\_UWS vs AdaBoost ensemble of 200 (depth = 2, forced to 4 terminal nodes) CART trees for 100 simulations. (*) in the test error indicates statistically significant ($\alpha = 0.05$) better performance on a paired t-test.}
		\begin{tabular}{lcccc}
			\toprule
			& \multicolumn{2}{c} {Test Error} & \multicolumn{2}{c} {Margin Improve} \\ [0.5ex] 
			Data Set & AdaBoost & UWS  & Mean & Min  \\
			\midrule
			Australian &0.0586*  & 0.0694 & 0.0100 & 0.0000\\
			BreastCancer &0.0152 & 0.0157  & 0.0700& 0.0000\\
			ColonCancer &0.1448* & 0.2103 &  0.1679  & 0.0137\\
			Diabetes  &0.1038*  & 0.1082&  0.0056 & 0.0000\\
			Four Class  &0.0497  & 0.0431*& 0.0176 & 0.0000\\
			Ionosphere  &0.0311*  & 0.0321 & 0.0389  & 0.0000\\
			Madelon  &0.0827 & 0.0827 & 0.0000  & 0.0000\\
			Mushrooms &0.0075*  & 0.0242& 0.0371  & 0.0000\\
			Musk &0.1046* & 0.1210 & 0.0348 & 0.0000\\
			Parkinsons &0.0440 & 0.0448 & 0.1190 & 0.0014\\
			Pima&0.2413* & 0.2489 & 0.0028 & 0.0000\\
			Sonar &0.0709*  & 0.0807 & 0.0925 & 0.0027\\
			Spambase & 0.0267* &0.0354& 0.0025& 0.0000\\
			Splice   &0.0750 & 0.0750 & 0.0000  & 0.0000\\	
			Transfusion&0.2261* & 0.2305 & 0.0015& 0.0000\\
			\bottomrule
		\end{tabular}%
		\label{tab:3}%
	\end{table}%
	
	\subsection{Margin Maximization (Decreasing Weights)}
	\label{Sec:9}
	
	As previously mentioned, Figure \ref{fig:1} shows the cumulative margin distributions for AdaBoost on the Splice data set for different values of $T$.  This is a typical behavior of AdaBoost similar to what \cite{schapire98} noted by saying that  ``boosting is especially aggressive at increasing the margins of the examples, so much so that it is willing to suffer significant reductions in the margins of those examples that already have large margins."  In this section, we test the hypothesis that a better performing ensemble should place more emphasis in the lower margins and deemphasize those observations with already high margins. Intuitively, this makes sense, as we want algorithms to place special focus on hard-to-classify observations. Here, we also modify Algorithm \ref{MM} and set $r_i = \left[(n+1) - rank(m_i)\right]^k$, which weights the margins according to their ranks in a nonlinear, decreasing way. This weighting scheme places more emphasis on the smaller margins, and places less weight on the larger margins. Empirical examples show that using higher values of $k$ results in more pronounced emphasis on the smaller margins, but the value of $k$ can be chosen via cross-validation or via trial and error depending on how pronounced we want the optimization of the smaller margins to be. The choice of $k$ also affects the performance differently depending on the complexity structure of the data set. We call this margin weighting approach the MM\_EWS (Exponential Weighting Scheme) algorithm. Figure \ref{fig:3} shows the optimized margins of the MM\_EWS method versus the original Random Forest margins  (with a 45-degree line added).  We also plot the two cumulative margin distributions (CMDs) for comparison. The example is based on a 300-tree Random Forest ensemble constructed on the Sonar data using $k = 5$.  Figure \ref{fig:3} illustrates how the optimized margins are at least as large as the original ensemble solution, and how the smallest percentiles have been especially emphasized. 
	\begin{figure*}
		\includegraphics[width=\textwidth]{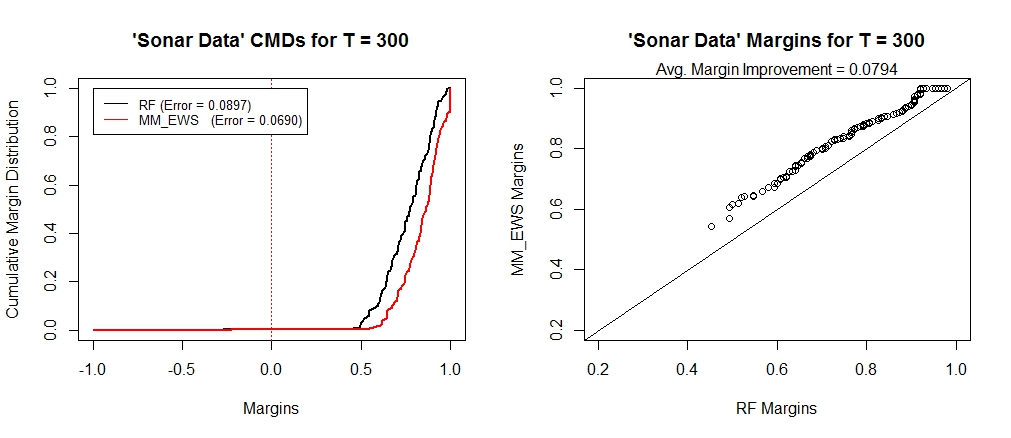}
		\caption{Comparisons of RF and the Exponential Weighting Scheme (MM\_EWS) with $k=5$ for the Sonar Data using 300 weak learners. The left panel compares the cumulative margin distributions (CMDs) for RF and the MM\_EWS algorithm. The right panel is a scatter plot of the AdaBoost margins and MM\_EWS margins. The line in the graph indicates equality of the margins.}
		\label{fig:3}       
	\end{figure*}
	We apply the MM\_EWS algorithm to Random Forests and AdaBoost solutions, arbitrarily setting $k = 5$. Table \ref{tab:4} illustrates the performance of the MM\_EWS algorithm versus Random Forests for 100 simulations using 200 decision trees. The results also indicate that there is no improved performance using this approach, in spite of the fact that the smaller margins are emphasized. Most of the experiments point to a better performance for the Random Forest solution, indicating that even though the margins were optimized and a special focus was put on the smaller margins, that in fact, can be detrimental to the prediction accuracy obtained by the original Random Forest solution. Table \ref{tab:5}, shows the performance of AdaBoost versus the MM\_EWS algorithm. The results also continue to hold here and are in fact very similar to those obtained using Random Forests.  Only in one of the data sets analyzed, the improved margins solutions results in a statistically better performing ensemble compared to the original AdaBoost solution.

	\begin{table}
		\centering
		\caption{MM\_EWS vs Random Forests ensemble of 200 (depth = 2, forced to 4 terminal nodes) CART trees for 100 simulations. (*) in the test error indicates statistically significant ($\alpha = 0.05$) better performance on a paired t-test.}
		\begin{tabular}{lcccc}
			\toprule
			& \multicolumn{2}{c} {Test Error} & \multicolumn{2}{c} {Margin Improve} \\ [0.5ex] 
			Data Set & RF & EWS  & Mean & Min  \\
			\midrule
			Australian &0.0534  & 0.0535 & 0.0128 & 0.0000\\
			BreastCancer &0.0115*  & 0.0137  & 0.0199 & 0.0000\\
			ColonCancer &0.0786* & 0.1079 &  0.1855 & 0.0325\\
			Diabetes  &0.1012  & 0.1012 & 0.0000 & 0.0000\\
			Four Class  &0.0021* & 0.0034 & 0.0274 & 0.0000\\
			Ionosphere  &0.0298  & 0.0306 & 0.0411 & 0.0000\\
			Madelon  &0.0775 & 0.0775 & 0.0000  & 0.0000\\
			Mushrooms &0.0000  & 0.0000 & 0.0021  & 0.0000\\
			Musk &0.0499 & 0.0475* & 0.0271 & 0.0014\\
			Parkinsons &0.0437 & 0.0426 & 0.0726 & 0.0000\\
			Pima&0.1012 & 0.1012 & 0.0000 & 0.0000\\
			Sonar & 0.0858  & 0.0823* & 0.0632 & 0.0087\\
			Spambase & 0.0209 &0.0209 & 0.0000& 0.0000\\
			Splice   &0.0659& 0.0659 & 0.0000  & 0.0000\\	
			Transfusion&0.1054* & 0.1065 & 0.0143& 0.0000\\
			\bottomrule
		\end{tabular}%
		\label{tab:4}%
	\end{table}%

	\begin{table}
		\centering
		\caption{MM\_EWS vs AdaBoost ensemble of 200 (depth = 2, forced to 4 terminal nodes) CART trees for 100 simulations. (*) in the test error indicates statistically significant ($\alpha = 0.05$) better performance on a paired t-test.}
		\begin{tabular}{lcccc}
			\toprule
			& \multicolumn{2}{c} {Test Error} & \multicolumn{2}{c} {Margin Improve} \\ [0.5ex] 
			Data Set & AdaBoost & EWS  & Mean & Min  \\
			\midrule
			Australian &0.0579*  & 0.0692 & 0.0114 & 0.0000\\
			BreastCancer &0.0173*  & 0.0233  & 0.0158 & 0.0000\\
			ColonCancer &0.1274 & 0.1294 &  0.0014  & 0.0000\\
			Diabetes  &0.1034*  & 0.1093&  0.0048 & 0.0000\\
			Four Class  &0.0498  & 0.0395* & 0.0274 & 0.0000\\
			Ionosphere  &0.0334*  & 0.0345& 0.0303 & 0.0002\\
			Madelon  & 0.1015* & 0.1285 & 0.0005  & 0.0000\\
			Mushrooms &0.0004  & 0.0002 & 0.1240  & 0.0000\\
			Musk &0.0663* & 0.0812& 0.0115& 0.0000\\
			Parkinsons &0.0389 & 0.0409 & 0.0967 & 0.0095\\
			Pima&0.1014 & 0.1014& 0.0000 & 0.0000\\
			Sonar &0.0754  & 0.0757 & 0.0698 &  0.0140\\
			Spambase & 0.0273* &0.0370& 0.0023& 0.0000\\
			Splice   &0.0750& 0.0750& 0.0000  & 0.0000\\	
			Transfusion&0.2261* & 0.2305 & 0.0015& 0.0000\\
			\bottomrule
		\end{tabular}%
		\label{tab:5}%
	\end{table}%

	\subsection{Margin Maximization (Lower Percentiles)}
	\label{Sec:9}
	\indent Here we focus on optimizing the lower margin percentiles. We also modify Algorithm \ref{MM} with the weights $r_i = I\{ \left[(n+1) - rank(m_i)\right]>(n-k)\}$, where $I\{ \}$ is the indicator function, to give equal weights (of 1) to the lowest $k$ margins, and zero weights to the remaining margins. The number of observation margins to optimize $k$ will be based on the percentage $\xi$  of lower margins to optimize, so that $k = \ceil{n\xi}$.  We call this approach the  MM\_PWS (Percentile Weighting Scheme) The main hypothesis we test here here is that improving examples with lower margins hold a better clue to the performance of ensembles than those with already large margins. 
	
	Figure \ref{fig:4} illustrates how the optimized margins of the MM\_PWS method compare versus the original Random Forest margins (with a 45-degree line added).  We also plot the two cumulative margin distributions (CMDs) for comparison. The example, based on a 300-tree Random Forest ensemble on the Sonar data, shows that although the new margins are at least as large as the original ensemble solution, the lowest $\xi=0.10$, corresponding to $k = 15$ margin instances have been especially emphasized. It is noteworthy in Figure \ref{fig:4} the fact that all of the margins have increased through the reweighting of the weak learner votes, but more so the lowest $k=15$. This produces a lower upper bound on the generalization error in (\ref{bound:1}), and should consequently produce a better performing ensemble solution. 
	\begin{figure*}
		\includegraphics[width=\textwidth]{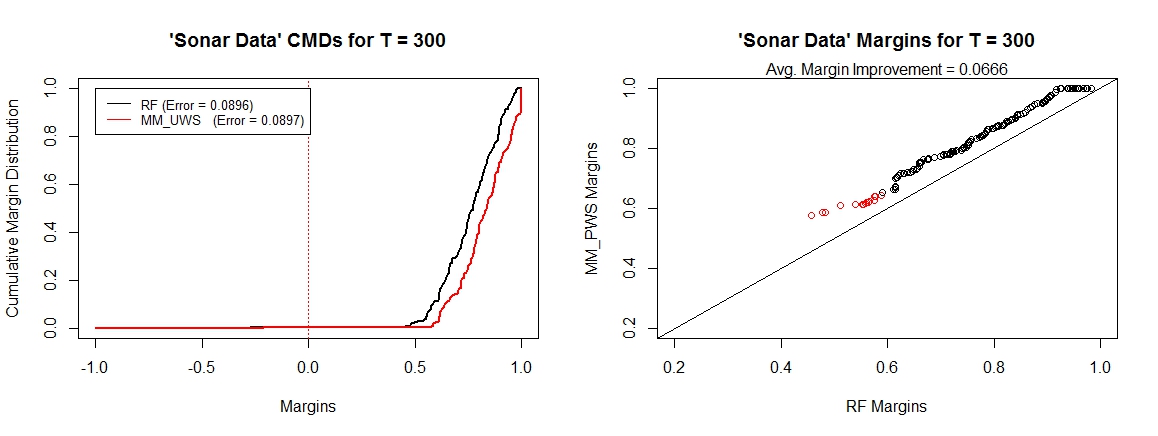}
		\caption{Comparisons of RF and the Percentile Weighting Scheme (MM\_PWS) with $\xi=0.10$ for the Sonar Data using 300 weak learners. The left panel compares the cumulative margin distributions (CMDs) for RF and the MM\_PWS algorithm. The right panel is a scatter plot of the AdaBoost margins and MM\_PWS margins. The line in the graph indicates equality of the margins. Red points indicate margins specifically optimized by the MM\_PWS method.}
		\label{fig:4}       
	\end{figure*}
	Tables \ref{tab:6} and \ref{tab:7} show the results on the performance of the MM\_PWS algorithm versus Random Forests (200 weak learners) and AdaBoost (200 weak learners) respectively for 100 simulations. Although the percentiles to optimize were set to $\xi=0.05, 0.20, 0.50$, the results still continue to hold for different values of $\xi$ not shown here. Noticeable from tables \ref{tab:6} and \ref{tab:7} is the fact that there is no improved performance in the test set errors using this approach either, in spite of the optimization algorithm specifically focusing on the lower margins. Most of the experiments instead point to a better performance for the original Random Forests and AdaBoost solutions, suggesting that optimizing the smaller margins does not yield any improvement and instead is a detrimental strategy for better performing ensemble algorithms.

	\begin{table}
		\centering
		\caption{MM\_PWS vs Random Forests ensemble of 200 (depth = 2, forced to 4 terminal nodes) CART trees for 100 simulations. (*) in the test error indicates statistically significant ($\alpha = 0.05$) better performance than the original RF solution and (-) indicates statistically significant ($\alpha = 0.05$) worse performance than the original RF solution.}
		\begin{tabular}{lcccc}
			\toprule
			& & \multicolumn{3}{c} {Test Set Error Rate} \\ [0.5ex] 
			Data Set & RF & PWS$_{.05}$  & PWS$_{.20}$ & PWS$_{.50}$  \\
			\midrule
			Australian &0.0554  & 0.0550 & 0.0553& 0.0553\\
			BreastCancer &0.0124  & 0.0143-  & 0.0146- & 0.0146- \\
			ColonCancer &0.0718 & 0.1009- &  0.0958- & 0.0958- \\
			Diabetes  &0.1023 & 0.1023 & 0.1023 & 0.1023\\
			Four Class  &0.0021 & 0.0037- & 0.0037- & 0.0037-\\
			Ionosphere  &0.0283  & 0.0284 & 0.0284 & 0.0284\\
			Madelon  &0.0814 & 0.0814 & 0.0814  & 0.0814\\
			Mushrooms &0.0000  & 0.0000 & 0.0000  & 0.0000\\
			Musk &0.0488 & 0.0484 &0.0489 &0.0489\\
			Parkinsons &0.0442 & 0.0418* & 0.0432 & 0.0432\\
			Pima&0.1012 & 0.1012 & 0.0000 & 0.0000\\
			Sonar & 0.0765  & 0.0751 & 0.0761 & 0.0761\\
			Spambase & 0.0209 &0.0209 & 0.0209& 0.0209\\
			Splice   &0.0651& 0.0651&0.0651  & 0.0651\\		
			Transfusion&0.1054 &  0.1064- &  0.1063- &  0.1063- \\
			\bottomrule
		\end{tabular}%
		\label{tab:6}%
	\end{table}%

	\begin{table}
		\centering
		\caption{MM\_PWS vs AdaBoost ensemble of 200 (depth = 2, forced to 4 terminal nodes) CART trees for 100 simulations. (*) in the test error indicates statistically significant ($\alpha = 0.05$) better performance than the original AdaBoost solution and (-) indicates statistically significant ($\alpha = 0.05$) worse performance than the original AdaBoost solution.}
		\begin{tabular}{lcccc}
			\toprule
			& & \multicolumn{3}{c} {Test Set Error Rate} \\ [0.5ex] 
			Data Set & AB & PWS$_{.05}$  & PWS$_{.20}$ & PWS$_{.50}$  \\
			\midrule
			Australian &0.0595  & 0.0687- & 0.0686- & 0.0687-\\
			BreastCancer &0.0149 & 0.0147 & 0.0147 & 0.0147\\
			ColonCancer &0.1270 & 0.1302- &  0.1302-  &0.1302-\\
			Diabetes  &0.1030  & 0.1080-&  0.1080- & 0.1080-\\
			Four Class  &0.0489  & 0.0411* & 0.0411* & 0.0411*\\
			Ionosphere  &0.0314  & 0.0327-& 0.0327- & 0.0327-\\
			Madelon  & 0.1015 & 0.1285- &0.1285-  & 0.1285-\\
			Mushrooms &0.0004  & 0.0000* & 0.0000*  & 0.0000*\\
			Musk &0.0670 & 0.0809-& 0.0809-& 0.0809-\\
			Parkinsons &0.0409 & 0.0418 & 0.0418 & 0.0418\\
			Pima&0.1027 & 0.1088-& 0.1088- & 0.1088-\\
			Sonar &0.0732  & 0.0785- & 0.0799- & 0.0815-\\
			Spambase & 0.0274 &0.0365-& 0.0365-& 0.0365-\\
			Splice   &0.0750& 0.0750&0.0750 & 0.0750\\	
			Transfusion&0.2261* & 0.2305 & 0.0015& 0.0000\\
			\bottomrule
		\end{tabular}%
		\label{tab:7}%
	\end{table}%
	
	\section{Reducing the Variation of the Margins}
	\label{maxvar}
	Hypotheses presented by \cite{reyzin2006boosting}, \cite{shen2010boosting} and \citet{germain2015risk} with Theorems 3 and 4 all suggest that reducing the variation of the margins might also improve the generalization performance of an ensemble. In this section, we devise optimization algorithms that focus on reducing the variation of the margins specifically. The idea of this approach is to ``squeeze" the margins either by directly reducing the variance and disregarding the complete distribution of the margins or by trying to push the mean of distribution while simultaneously decreasing the variance. We propose two methods to increase the lower percentiles and decrease the upper percentiles of the margin distribution for a given ensemble solution, or that directly reduce the variation of the margins, while maintaining or improving the whole margin distribution. We refer to this as ``squeezing the margins." 
	
	\subsection{Range Reduction Method (SM1)}
	\label{Sec:13}
	
	We propose here raising a proportion $\xi$ of the smallest margins to at least $\theta_\xi$, the $\xi$-th percentile of the current margin instances, while allowing margins above the current $\bar{m}$ to decrease, if necessary, but no lower than $\bar{m}$. Margins below $\bar{m}$ are required to be at least as large as $\theta_\xi$. We call this algorithm SM1 (Squeezing Method 1). The linear programming formulation for this problem is given in Algorithm \ref{SM1}.
	\begin{algorithm}
		\caption{SM1 Algorithm}
		\label{SM1}
		\[ \begin{array}{rl}
		\max  & \sum{_{i=1}^n\left[y_i\sum{_{t=1}^Tw_th_{it}-m_i}\right]} \\
		\mbox{s.t.} & y_i \sum{_{t=1}^T w_t  h_{it} \geq \theta_\xi} \text{, for $i$ such that $m_i \leq \bar{m}$} \\
		& y_i \sum{_{t=1}^T w_t  h_{it} \geq \bar{m}} \text{, for $i$ such that $m_i > \bar{m}$} \\
		& \sum{_{t=1}^T w_t = 1} \\
		&  w_t \geq 0, t =1,2,..., T \\
		\end{array}
		\]
	\end{algorithm} 
	\noindent The value of $\xi$ is selected by the user, and could be optimized through trial-and-error or cross-validation, but we have set $\xi=0.05$ for all simulations, as this does not impose an unrealistic burden to the optimization formulation. Setting $\xi$ too large might result in no feasible solution for the LP. Figure \ref{fig:5} shows the improved margins of the SM1 algorithm versus the original margins obtained by the Random Forest ensemble (with a 45-degree line added).  We also plot the two cumulative margin distributions (CMDs) for comparison. The example is based on a 300-tree Random Forest ensemble on the Sonar data, using $\xi=0.20$ for a more visual understanding of how the algorithm works. Note that the margins  between $\theta_\xi$ and $\bar{m}$ generated by the SM1 algorithm do not necessarily have to be greater than those of the original ensemble solution, and new margins greater than $\bar{m}$ can also be lower than the ensemble solution, as long as they are larger than $\bar{m}$. In Figure \ref{fig:5} all of the margins have increased and are at least as large as $\theta_\xi = 0.6467$, which corresponds to the 20\% percentile of the margins  for the RF solution. It is also noteworthy to mention that the particular LP formulation in the SM1 algorithm attempts to minimize the range of the margins, and the variance of the margins might or might not be lower than that of the original ensemble solution. For the example in Figure \ref{fig:5} both the variance and the range were lower than those resulting from the original Random Forest solution, with a reduction of 0.0012 and 0.1733 on the variance and range respectively, however the performance does not show any improvement and in fact the opposite happens here.
	\begin{figure*}
		\includegraphics[width=\textwidth]{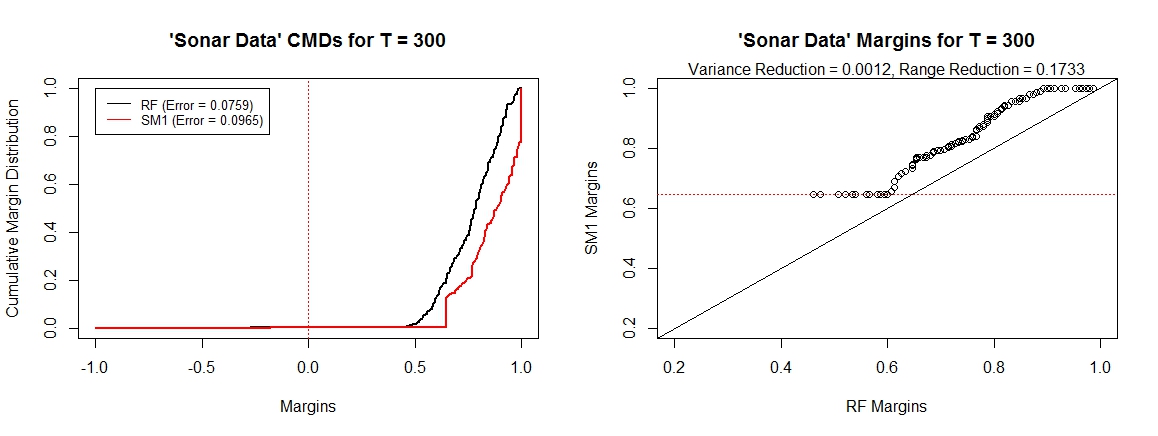}
		\caption{Comparisons of RF and the Squeezing Method 1 (SM1) with $\xi=0.20$ for the Sonar Data using 300 weak learners. The left panel compares the cumulative margin distributions (CMD’s) for RF and the SM1 algorithm. The right panel is a scatter plot of the AdaBoost margins and SM1 margins. The line in the graph indicates equality of the margins. Red points indicate margins specifically optimized by the SM1 method.}
		\label{fig:5}       
	\end{figure*}
	Tables \ref{tab:8} and \ref{tab:9} display the test set error rates for the proposed method vs Random Forests and AdaBoost, respectively, along with the reduction in variance and range of the margin distribution of the optimized margins. The results are based on 100 simulations using 200 decision trees of depth 2.  The fact that, more often than not, the performance of the proposed method is worse than that of the original ensembles suggests that the strategy of reducing the range of the margin distribution of an ensemble is not optimal.

	\begin{table}
		\centering
		\caption{SM1 vs Random Forests ensemble of 200 (depth = 2, forced to 4 terminal nodes) CART trees for 100 simulations. (*) in the test error indicates statistically significant ($\alpha = 0.05$) better performance on a paired t-test.}
		\begin{tabular}{lcccc}
			\toprule
			& \multicolumn{2}{c} {Test Error} & \multicolumn{2}{c} {Reduction} \\ [0.5ex] 
			Data Set & RF & SM1$_{0.05}$  & Var & Range  \\
			\midrule
			Australian &0.0559  & 0.0561 & 0.0033 & 0.2687\\
			BreastCancer &0.0120*  & 0.0123  & 0.0022 & 0.1417\\
			ColonCancer &0.0849* & 0.1105 &  0.0094 & 0.3298\\
			Diabetes  &0.1012*  & 0.1036 & 0.0008 & 0.2026\\
			Four Class  &0.0020* & 0.0030 & 0.0033 & 0.4582\\
			Ionosphere  &0.0268*  & 0.0280 & 0.0076 & 0.3151\\
			Madelon  &0.0775 & 0.0775 & 0.0000  & 0.0000\\
			Mushrooms &0.0000  & 0.0000 & 0.0004  & 0.2904\\
			Musk &0.0478* & 0.0495 & -0.0006 & 0.2106\\
			Parkinsons &0.0426 & 0.0455 & 0.0090 & 0.2254\\
			Pima&0.1024* & 0.1047 & 0.0012 & 0.2023\\
			Sonar & 0.0780  & 0.0877 & -0.0060 & 0.1045\\
			Spambase & 0.0212 &0.0212 & 0.0000& 0.0000\\
			Splice   &0.0652*& 0.0667 & 0.0006  & 0.0690\\	
			Transfusion&0.1055 & 0.1055 & 0.0000& 0.0000\\
			\bottomrule
		\end{tabular}%
		\label{tab:8}%
	\end{table}%

	\begin{table}
		\centering
		\caption{SM1 vs AdaBoost ensemble of 200 (depth = 2, forced to 4 terminal nodes) CART trees for 100 simulations. (*) in the test error indicates statistically significant ($\alpha = 0.05$) better performance on a paired t-test.}
		\begin{tabular}{lcccc}
			\\
			\toprule
			& \multicolumn{2}{c} {Test Error} & \multicolumn{2}{c} {Reduction} \\ [0.5ex] 
			Data Set & RF & SM1$_{0.05}$   & Var & Range  \\
			\midrule
			Australian &0.0590*  & 0.06106& 0.0052& 0.0782\\
			BreastCancer &0.0144  & 0.0149 & -0.0007 & 0.0065\\
			ColonCancer &0.0786* & 0.1079 &  0.1855 & 0.0325\\
			Diabetes  &0.1012*  & 0.1108 & 0.0050 & 0.2198\\
			Four Class  &0.0469* & 0.0596 & -0.0025& 0.0744\\
			Ionosphere  &0.0307* & 0.0352 & -0.0141 & 0.1331\\
			Madelon  &0.0775 & 0.0775 & 0.0000  & 0.0000\\
			Mushrooms &0.0004  & 0.0002* & -0.0084  & 0.2261\\
			Musk &0.0644 & 0.0798 & -0.0127 & 0.1009\\
			Parkinsons &0.0437 & 0.0426 & 0.0726 & 0.0000\\
			Pima&0.1028* & 0.1108 & 0.0043 & 0.1990\\
			Sonar & 0.0741*  & 0.0888 & -0.0299& 0.1098\\
			Spambase & 0.0268* &0.0420 & -0.0026& 0.4633\\
			Splice   &0.0659& 0.0659 & 0.0000  & 0.0000\\	
			Transfusion&0.0964* & 0.1068 & 0.0021& 0.2717\\
			\bottomrule
		\end{tabular}%
		\label{tab:9}%
	\end{table}%

	\subsection{Regression-Based Variance Reduction (SM2)}
	\label{Sec:14}
	
	We propose a regression-based approach to reduce the variation in the margins of a given ensemble solution. With a little creativity, we can express our optimization of margins problem as a regression fitting problem, where the response is the average margins $\bar{m}$ and the covariate matrix is given by $H$ in (\ref{eq:3}). To further clarify the proposed method, we define the following: \\
	
	\indent $y_i \equiv \bar{m}$ \indent \indent \text{for} $ i = 1,2,...,n$  \\
	\indent $x_{it} \equiv y_ih_{it}$ \indent \text{for} $ i = 1,2,...,n;  t = 1,2,...,T$  \\
	\indent $b_j \equiv w_t$  \indent  \indent \text{regresion through the origin} \\
	\indent $b_0 \equiv 0$  \indent \indent \text{ no intercept} 
	\\
	
	\noindent 
	With OLS regression we are trying to minimize the sum of squared residuals (SSE), which is given by $SSE = \sum_{i=1}^n   (y_i - \sum_{j=1}^pb_jx_{ij})^2$. This in turn translates into:
	\begin{equation}
	\label{eq:5}
	SSE = \sum_{i=1}^n (\bar{m} - y_i \sum_{t=1}^T\alpha_th_{it})^2 = \sum_{i=1}^n (m_i - \bar{m})^2,
	\end{equation}

	\noindent which is equivalent to minimizing the variance of the margins (around any
	value $\bar{m}$, which no longer has to be the mean of the margins). We can increase the value of $\bar{m}$  to trade off increasing the mean and decreasing the variance of the margin distribution. We can also use alternative regression criteria, including LTS (least trimmed sum of squares), LAV, LMS, Chebyshev (which should be equivalent to our SM2 solution), or any other robust/resistant form of regression (i.e., measure of variation).
	\begin{figure*}
		\includegraphics[width=\textwidth]{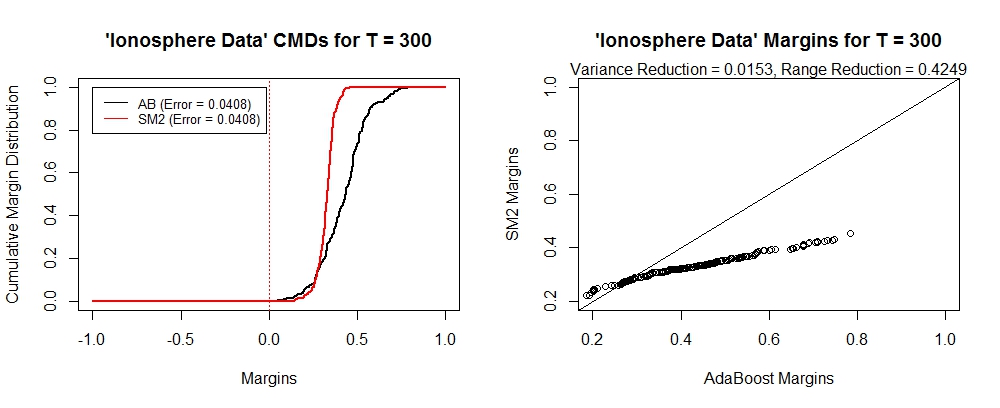}
		\caption{Comparisons of AdaBoost and the Regression-Based Squeezing Method (SM2) for the Ionosphere Data using 300 weak learners. The left panel compares the cumulative margin distributions (CMD’s) for AdaBoost and the SM2 algorithm. The right panel is a scatter plot of the AdaBoost margins and SM2 margins. The line in the graph indicates equality of the margins.}
		\label{fig:6}       
	\end{figure*}
	Figure \ref{fig:6} illustrates the results of the SM2 algorithm versus the original margins (with a 45-degree line added) for an AdaBoost solution of 300 (4-node, depth = 2) trees.  We also plot the two cumulative margin distributions (CMDs) for comparison. For this particular example the variance of the margins has been reduced by 0.0153, while the range has been reduced by 0.4249. There are three main issues to take into account when using the proposed approach: (1) the margins are no longer restricted to the
	interval [{1, +1], (2) the sum of the weights is not necessarily equal to 1, and (3) the weights can be negative. The best solution to these issues can be to (1) relax the assumption of non-negative weights to allow for negative weights, and (2) normalize the weights so they sum to zero. 
		
		In regression through the origin, if the $y$ values are multiplied by a constant, then the new regression coefficients are the original regression coefficients multiplied by the same constant. For simple linear regression through the origin, the least squares slope estimate is given by $b = \left( \sum_{i=1}^nx_iy_i\right) / \left( \sum_{i=1}^nx_i^2 \right)$, which verifies the result for simple linear regression. The biggest problem with allowing negative weights is that the margins are no longer bounded in the interval [{-1, +1]. At first, we thought this was a problem, but it does not appear that this is a necessary restriction. We still have to maintain the restriction of the weights summing to 1. By analogy, stock portfolio optimization in finance allows for negative weights with these same restrictions, so negative weights may not be a problem in this situation.

			It is clear from Tables \ref{tab:10} and \ref{tab:11} that reducing the variance of the margins does not necessarily result in a better performing ensemble. The results indicate that minimizing the variance of the margins from an AdaBoost or Random Forest solution increases the test set error rates more often than not for a RF ensemble even when the entire margin distribution has increased. For the AdaBoost ensemble, the results are inconclusive, as the SM2 algorithm performs better in 4 out of the 15 data sets, while AdaBoost performs better in 5. 
			
			\begin{table}
				\centering
				\caption{SM2 vs Random Forests ensemble of 200 (depth = 2, forced to 4 terminal nodes) CART trees for 100 simulations. (*) in the test error indicates statistically significant ($\alpha = 0.05$) better performance on a paired t-test.}
				\begin{tabular}{lcccc}
					\toprule
					& \multicolumn{2}{c} {Test Error} & \multicolumn{2}{c} {Reduction} \\ [0.5ex] 
					Data Set & RF & SM2  & Var & Range  \\
					\midrule
					Australian &0.0566*  & 0.0634 & 0.0228 & 0.3136\\
					BreastCancer &0.0122*  &0.0519  & 0.0126 & 0.7871\\
					ColonCancer &0.0779* & 0.1800 &  0.0210 & 0.5729\\
					Diabetes  &0.1030*  & 0.1111 & 0.0157 & 0.0882\\
					Four Class  &0.0023* &0.0048 & 0.0045 & 0.5067\\
					Ionosphere  &0.0279* & 0.1048 & 0.0191 & 0.7076\\
					Madelon  &0.0775 & 0.0775 & 0.0000  & 0.0000\\
					Mushrooms &0.0000  & 0.0000 & 0.0004  & 0.2950\\
					Musk &0.0494* & 0.0554& 0.0126 & 0.2810\\
					Parkinsons &0.0450* & 0.1169& 0.0246 & 0.6835\\
					Pima&0.1020* & 0.1111& 0.0159 & 0.0823\\
					Sonar & 0.0730  &0.1909& 0.0174 & 0.5741\\
					Spambase & 0.0211 &0.0216& 0.0059&0.0435\\
					Splice   &0.0652*& 0.0667 & 0.0006  & 0.0690\\	
					Transfusion&0.1053 & 0.1235 & 0.0281& -0.4832\\
					\bottomrule
				\end{tabular}%
				\label{tab:10}%
			\end{table}%

			\begin{table}
				\centering
				\caption{SM2 vs AdaBoost ensemble of 200 (depth = 2, forced to 4 terminal nodes) CART trees for 100 simulations. (*) in the test error indicates statistically significant ($\alpha = 0.05$) better performance on a paired t-test.}
				\begin{tabular}{lcccc}
					\toprule
					& \multicolumn{2}{c} {Test Error} & \multicolumn{2}{c} {Reduction} \\ [0.5ex] 
					Data Set & RF & SM2  & Var & Range  \\
					\midrule
					Australian &0.0631   & 0.0623& 0.0434&  0.3569\\
					BreastCancer &0.0151  & 0.0187 & 0.0173 & -0.3713\\
					ColonCancer &0.0786* & 0.1079 &  0.1855 & 0.0325\\
					Diabetes  &0.1021*  & 0.1066 & 0.0299 & 0.4157\\
					Four Class  &0.0484 & 0.0245* & 0.0039& -1.2648\\
					Ionosphere  &0.0319* & 0.0436 & 0.0198 &  0.3480\\
					Madelon  &0.0775 & 0.0775 & 0.0000  & 0.0000\\
					Mushrooms &0.0004*  & 0.0045& 0.0165  & -0.0043\\
					Musk &0.0500* & 0.0581 & 0.0125 & 0.2823\\
					Parkinsons &0.0625 & 0.0511*& 0.0489 & 0.3056\\
					Pima& 0.1012 & 0.1006 & 0.0150 & 0.0967\\
					Sonar & 0.0879  & 0.0853 & 0.0141& 0.2001\\
					Spambase & 0.0321 &0.0296* & 0.0175& 0.1741\\
					Splice   &0.1780& 0.1890 & 0.0111 & 0.0653\\	
					Transfusion&0.1010 & 0.0994* & 0.0372& 0.1537\\
					\bottomrule
				\end{tabular}%
				\label{tab:11}%
			\end{table}%

\section{Discussion and Future Research}
\label{conc}
\indent \indent The current state of research in explaining why ensemble methods perform as well as they do suggests that margins play a pivotal role. Upper bounds on the generalization error in (\ref{bound:1}) developed by \cite{schapire98} led to the ``large margins theory," which suggests that, holding other factors fixed such as the complexity of the base learning algorithm, larger margins should lead to better generalization performance.  \cite{breiman1999prediction} pointed out with his arc-gv algorithm that maximizing the minimum margin was not an optimal strategy. In fact, \cite{breiman1999prediction} also found in his experiments that arc-gv not only produced larger minimum margins over all training examples, but it also produced better margin distributions than those of AdaBoost, with generally worse test set error performance.  \cite{reyzin2006boosting} replicated the analysis in \cite{breiman1999prediction}, but pointed out that the trees (weak learners) generated by arc-gv were deeper on average than the trees (weak learners) generated by AdaBoost, which by (\ref{bound:1}) resulted in more complex trees, and hence the explanation of why arc-gv had a worse generalization performance. Many other researchers have developed tighter bounds and have reinforced the results in \cite{schapire98}, indicating that maximizing the margin distribution should result in better performing ensemble \citep{mason2000improved, shen2010boosting, wang2011refined, gao2013doubt}. In this paper, we have developed algorithms especifically designed to test whether the hypotheses based on the large margins theory hold. To avoid the mishaps found in \cite{breiman1999prediction}, we have fixed the complexity of the base learning algorithm by building CART trees forced to depth 2 (4-node decision trees). In our experiments, we have tested the main ideas on the current state of research in explaining ensemble performance with margins and found that they do not improve the generalization performance of the ensembles studied here, and most of the time, the performance worsens in fact. Although we do not rule out the importance of margins, and do not question the theory and mathematical underpinnings on the bounds, we believe the results presented here indicate that there might be other factors influencing the generalization performance of ensembles.  \cite{breiman1999prediction} concluded that given how loose the bounds are, this ``casts doubt on the ability of the loose VC-type bounds to uncover the mechanism leading to low generalization error." We hope that this study encourages more researchers to look beyond the current explanation of ensemble performance, in an effort to better explain and design superior performing ensemble algorithms.


\end{document}